\documentclass[conference]{IEEEtran}
\IEEEoverridecommandlockouts
\usepackage{cite}
\usepackage{amsmath,amssymb,amsfonts}
\usepackage{algorithmic}
\usepackage{graphicx}
\usepackage{textcomp}
\usepackage{xcolor}
\usepackage{tabularx}
\usepackage{diagbox}
\usepackage{subfig}
\usepackage{float}
\usepackage{multirow}
\usepackage{mathtools}
\usepackage{amsmath}
\usepackage{tikz}
\usepackage{hyperref}
\usepackage{url}
\usepackage{verbatim} 
\makeatletter

\def\ps@IEEEtitlepagestyle{
  \def\@oddfoot{\mycopyrightnotice}
  \def\@evenfoot{}
}
\def\mycopyrightnotice{
  {\footnotesize xxx-x-xxxx-xxxx-x/xx/\$31.00~\copyright~2018 IEEE\hfill} % <--- Change here
  \gdef\mycopyrightnotice{}
}

\@ifundefined{showcaptionsetup}{}{
 \PassOptionsToPackage{caption=false}{subfig}}
\usepackage{subfig}
\makeatother

\usepackage{eso-pic}
\newcommand\AtPageUpperMyright[1]{\AtPageUpperLeft{
 \put(\LenToUnit{0.5\paperwidth},\LenToUnit{-1cm}){
     \parbox{0.5\textwidth}{\raggedleft\fontsize{9}{11}\selectfont #1}}
 }}
\newcommand{\conf}[1]{
\AddToShipoutPictureBG*{
\AtPageUpperMyright{#1}
}
}

\conf{This Article has been accepted in  17th IEEE International Conference on Service Operations and Logistics, and Informatics (SOLI 2023) 11 – 13 December 2023, Singapore.} % Change according to their suggestion

\def\BibTeX{{\rm B\kern-.05em{\sc i\kern-.025em b}\kern-.08em
    T\kern-.1667em\lower.7ex\hbox{E}\kern-.125emX}}

\begin{document}

\title{Active SLAM Utility Function Exploiting Path Entropy\\

\thanks{*This work was carried out in the framework of the NExT Senior Talent Chair DeepCoSLAM, which was funded by the French Government, through the program Investments for the Future managed by the National Agency for Research ANR-16-IDEX-0007, and with the support of Région Pays de la Loire and Nantes Métropole.}
}

\author{\IEEEauthorblockN{Muhammad Farhan Ahmed}
\IEEEauthorblockA{\textit{LS2N} \\
\textit{Ecole Centrale de Nantes (ECN)}\\
Nantes, France \\
Muhammad.Ahmed@ec-nantes.fr}
\and
\IEEEauthorblockN{Vincent Frémont}
\IEEEauthorblockA{\textit{LS2N} \\
\textit{Ecole Centrale de Nantes (ECN)}\\
Nantes, France \\
vincent.fremont@ec-nantes.fr}
\and
\IEEEauthorblockN{Isabelle Fantoni}
\IEEEauthorblockA{\textit{LS2N} \\
\textit{CNRS}\\
Nantes, France \\
isabelle.fantoni@ls2n.fr}

}

\maketitle

\begin{abstract}
In this article we present a utility function for Active SLAM (A-SLAM) which utilizes map entropy along with D-Optimality criterion metrices for weighting goal frontier candidates. We propose a utility function for frontier goal selection that exploits the occupancy grid map by utilizing the path entropy and favors unknown map locations for maximum area coverage while maintaining a low localization and mapping uncertainties. We quantify the efficiency of our method using various graph connectivity matrices and map efficiency indexes for an environment exploration task. Using simulation and experimental results against similar approaches we achieve an average of 32\% more coverage using publicly available data sets.
\end{abstract}


%%%%%%%%%%%%%%%%%%%%%%%%%%%%%%%%%%%%%%%%%%%%%%%%%%%%%%%%%%%%%%%%%%%%%%%%%%%%%%%%
\section{INTRODUCTION}

Simultaneous Localization and Mapping (SLAM) encompasses a range of methods used by robots to determine their own position while simultaneously creating a map of their surroundings during navigation. Most SLAM algorithms are considered passive, wherein the robot is moving freely, and the navigation or path-planning algorithm plays no active role in directing the robot's movements or trajectory. In contrast, A-SLAM aims to address the optimal exploration of unknown environments by proposing a navigation strategy that generates future target positions and actions. These actions are designed to reduce uncertainty in the map and the robot's pose, enabling a fully autonomous SLAM system.

In A-SLAM, the robot's exploration process begins by initially identifying potential target positions within its current map estimate. Once the robot has established a map of its surroundings, it proceeds to locate positions worth exploring. One commonly used method for this task is frontier-based exploration, initially introduced by \cite{yamuchi}. Frontier-based exploration defines the 'frontier' as the boundary separating known map locations from unknown ones, as observed by the robot's sensors. After identifying these goal frontiers, the robot computes a cost or utility function. This function relies on the potential reward associated with selecting the optimal action from a set of all possible actions. In an ideal scenario, this utility function should account for the complete joint probability distribution of both the map and the robot's poses. To quantify this uncertainty, we turn to two well-established domains: Information Theory (IT), and the Theory of Optimal Experimental Design (TOED), as detailed in \cite{S23}. The subsequent crucial step involves executing the optimal action, guiding the robot towards its goal position using path planning techniques.

 In this article, we propose a utility function for selecting the goal frontier candidate for autonomous exploration by the robot. Our function takes into account the amount of uncertainty in the map measured as path entropy and Euclidean distance to each frontier candidate. We add this utility function to the one of \cite{NP4} which selects the frontier weighted by the D-optimality criterion as the maximum number of spanning trees (in pose-graph) towards it. Using the proposed utility function brings the advantage that it incorporates not only the SLAM uncertainty but also entropy reduction in the environment to guide the robot to promising unknown areas for the exploration task. 

This article is organized as follows: Section \ref{RELATED WORK}, provides an insight into the related work. Section \ref{PRELIMINARIES} gives preliminary knowledge about the structure of modern graph SLAM, A-SLAM, and how uncertainty is measured and related to graph connectivity. In Section \ref{METHODOLOGY} we present the approach and formulation of our proposed method. Section \ref{SIMULATION RESULTS} presents and discusses our simulation results. Finally, we conclude in Section \ref{CONCLUSIONS} summarizing our contributions and motivating future research aspects.

\section{RELATED WORK}\label{RELATED WORK}
As discussed above, A-SLAM involves frontier detection, utility computation by quantifying and minimizing the uncertainty, and finally generating the action for robot navigation. 

 The method proposed by \cite{AS10} formulates a hybrid control switching exploration method. Within the occupancy grid map, each frontier is segmented, a trajectory is planned for each segment, and the trajectory with the highest map segment covariance is selected from the global cost map, which renders this method computationally expensive and limited to static obstacles. Meanwhile, the approach in \cite{AS23} deals with dynamic environments with multiple ground robots and uses frontier exploration for autonomous exploration, and a utility function based on Shannon and Renyi entropy \cite{entropy} is used for the computation of the utility of paths. 

When dealing with uncertainty quantification from  IT perspective, the authors of \cite{AS9} address the joint entropy minimization exploration problem and propose two modified versions of Rapidly exploring Random Trees (RRT) \cite{S3}, as dRRT* and eRRT* respectively. dRRT* uses distance, while eRRT* uses entropy change per distance traveled as the cost function. Actions are computed in terms of the joint entropy change per distance traveled. The simulation results proved that a combination of both of these approaches provides the best path-planning strategy. An interesting comparison between IT approaches is given in \cite{AS7} where frontier-based exploration is deployed to select future candidate target positions. A comparison of joint Entropy reduction between the robot path and map is done against Expected Map Mean (EMM) and Kullback-Leiber Divergence. It was concluded that most of these approaches were not able to properly address the probabilistic aspects of the problem and are most likely to fail because of high computational cost and map grid resolution dependency on performance. The authors in \cite{AS21} use entropy reduction only over map features and use an entropy metric based on Laplacian approximation with a unified quantification of exploration and exploitation gains.

Recently in the works of \cite{NP7} and \cite{NP9}, the authors exploit the graph SLAM connectivity and pose it as an Estimation over Graph (EoG) problem, where each node (state vector) and vertex (measurement) connectivity is strongly related to the SLAM estimation reliability. By exploiting the spectral graph theory which deals with the Eigenvalues, Laplacian, and Degree matrix of the associated Fisher Information Matrix (FIM) and graph connectivity respectively the authors state that the graph Laplacian is related to the SLAM information matrix and the number of Weighted number of Spanning trees (WST) is directly related to the estimation accuracy in graph SLAM.

The authors in \cite{NP1}\cite{NP4} extend \cite{NP9} by debating that the maximum number of WST is directly related to the Maximum Likelihood Estimate (MLE) of the underlying graph SLAM problem formulated over lie algebra. Instead of computing the D-optimality criterion over the entire SLAM sparse information matrix, a modern D-optimality criterion is computed over the weighted graph Laplacian where each vertex is weighted using edge D-Optimality. Furthermore, it is proven that the maximum number of WST of this weighted graph Laplacian is directly related to the underlying pose graph uncertainty.

\section{PRELIMINARIES}\label{PRELIMINARIES}
\subsection{Graph SLAM}\label{Graph SLAM}
\begin{figure}[thpb]
    \centering
    \includegraphics[height=2.5cm ,width=7.5cm]{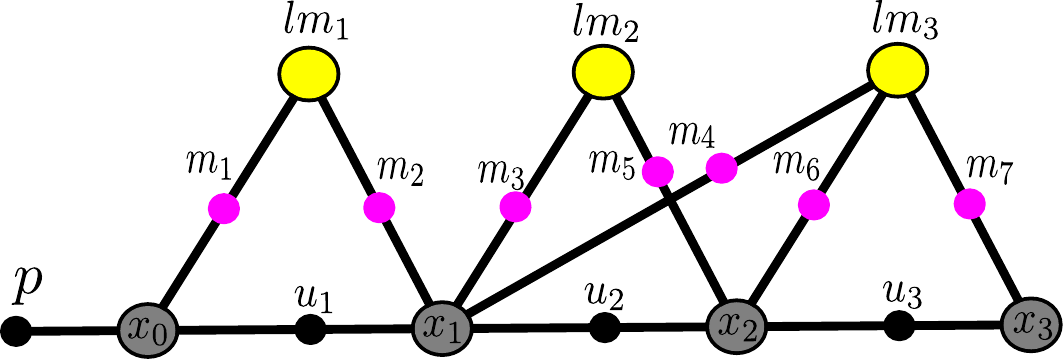}  
    \caption{Graph SLAM structure}
    \label{fig:0}
\end{figure}

Modern SLAM approaches adopt a graphical approach (bipartite graph) where each node represents the robot or landmark pose and each edge represents a pose to pose or pose to landmark measurement. As an example for a robot with four pose states,  shown in Fig. \ref{fig:0} $x_{0:3}$, $lm_{1:3}$ $u_{1:3}$, $m_{1:7}$ represent the robot pose, landmark pose, robot pose measurement, and landmark measurement respectively. The objective is to find the optimal state vector $x^*$ which minimizes the measurement error $e_i(x)$ weighted by the covariance matrix $\Omega_i \in \mathbb{R}^{l \times l}$ where $l$ is the dimension of the state vector $x$ as shown in Equation \ref{slam:eq1}. We direct interested readers to \cite{graphslam} for an introduction and comparison of SLAM methods. 
 
\begin{comment}
  
\end{comment}

 \begin{equation} 
\label{slam:eq1}
	\mathit{x}^{*} = \arg \min_{x} \sum_{i} \mathbf{e}_i^{T}(\mathit{x})\Omega_i\mathbf{e}_i(\mathit{x})
\end{equation}

In A-SLAM the robot has to navigate in an unknown environment by performing actions in the presence of noisy sensor measurements that reduce its state and map uncertainties with respect to the environment. Such a scenario is modeled as an instance of the Partially Observable Markov Decision Process (POMDP) \cite{thurn}  \cite{NP3}\cite{NP5}. 
\begin{comment}
POMDP is defined as a 7 tuple  $(X,A,O,T,\rho_{o},\beta,\gamma)$, where $X \in \mathbb{R}$ represents the robots state space and is represented as the current state $x\in X$ and next state $x^{'}\in X$, $A \in \mathbb{R}$ is the action space and can be expressed as $a\in A$, $O$ are the observations where $o\in O$, $T$ is the state transition function between an action ($a$), present state ($x$) and next state ($x^{'}$), $T$ accounts for robot control uncertainty in reaching the new state $x^{'}$, $\rho_{o}$ accounts for sensing uncertainty, $\beta$ is the reward associated with the action taken in state $x$, $\gamma \in (0,1)$ takes into account the discount factor ensuring a finite reward even if the planning task has an infinite horizon. Both $T(x,a,x^{'}) = p(x^{'}\mid x,a)$ and $\rho_{o}(x,a,o) = p(o\mid x^{'},a)$ can be expressed using conditional probabilities.   
\end{comment}

Let us consider a scenario where the robot is in state $x$ and takes an action $a$ to move to $x^{’}$. The robot’s goal is to choose the optimal policy $\chi^{*}$ that maximizes the associated expected reward ($\mathbb{E}$) for each state-action pair and it can be modeled as Equation \ref{eqn:3}:
\begin{equation}
\label{eqn:3}
    \chi^{*} =\operatorname*{argmax}_t \sum_{t=0}^{\infty} \mathbb{E}\Gamma^{t}\alpha(x_t,a_t)
\end{equation}

Where $x_t$, $a_t$ and $\Gamma^{t}$ are respectively the state, action, and discount factor evolution at time $t$ and $\alpha(x_t,a_t)$ is the reward function associated with the action performed to reach $x$. Although the POMDP formulation of A-SLAM is the most widely used approach, it is considered computationally expensive as it considers planning and decision-making under uncertainty. For computational convenience, it is divided into three main steps which identify the potential goal positions/waypoints, compute the cost to reach them, and then select actions based on a utility criterion that decreases map uncertainty and increases the robot’s localization.

\subsection{Uncertainty Quantification}\label{Uncertainty Quantification}
The cost or utility is computed based on the reward value of the optimal action selected from a set of all possible actions according to Equation \ref{eqn:3}. For uncertainty quantification IT and TOED methods are used.

In IT, Shannon entropy/entropy measures the amount of uncertainty associated with a random variable or random quantity. Since robot pose and the map is estimated as a multivariate Gaussian distributions the authors in \cite{stachniss} describe Shannon entropy of the map $\textit{E} \in (0,1)$  as in Equation \ref{eq:3} where the map $\text{\textit{M}}$ is represented as an occupancy grid and each cell $c_{i,j}$ is associated with a probabilistic estimation $P(c_{i,j})$ of its value of $1$ being occupied and $0$ as free. The objective is to reduce both the robot pose and map entropy. However, measuring entropy can be computationally demanding because computing probabilistic estimations of both the robot's position and the map over the entire map area is required along with its associated grid resolution.

\begin{multline}
\label{eq:3}
	\textit{E}[p(\textit{M})] = - \sum_{i,j}(p(c_{i,j})log(p(c_{i,j})) \\ + (1-p(c_{i,j}))log(1-p(c_{i,j}))
\end{multline}

 Alternatively, if we consider task driven utility functions, the uncertainty metric is evaluated by reasoning over the propagation of uncertainty associated with the Fisher Information Matrix (FIM) $\Delta = \Omega^{-1}$ of graph SLAM. TOED provides many optimally criteria, which give a mapping of the covariance matrix to a scalar value. Less covariance contributes to a higher weight of the action set $\chi$. For a covariance matrix $\Omega \in \mathbb{R}^{n \times n}$ and having eigenvalues $\zeta_n$, A,D and E optimality criterion are defined which minimize the average variance, covariance ellipsoid and maximum Eigenvalue.

\begin{comment}

\begin{equation}
\label{eqn:7}
	D-Opt\overset{\Delta}{=}exp(\frac{1}{n}\sum_{k=1}^{n}log(\zeta_k))
\end{equation}
    
\end{comment} 

TOED approaches require both the robot pose and map uncertainties to be represented as covariance matrix and may be computationally expensive, especially in landmark-based SLAM where its size increases as new landmarks are discovered. Hence, IT-based approaches are preferred over TOED.

In \cite{NP7} and \cite{NP9} the authors debate how the graphical topology of SLAM  has an impact on estimation reliability. They establish a relationship between the Weighted number of Spanning Trees (WST) and the D-Optimality criterion and show the graph Laplacian is closely related to the FIM. Three graph connectivity metrics are discussed with connection to the SLAM estimation accuracy, namely: 1) Algebraic Connectivity (A.C), defines the robustness/resilience of a graph to stay connected even after the removal of some nodes. For connected undirected graph $\mathit{G}=(\mathit{v},\mathit{\epsilon})$, having $\mathit{v}$ vertices (poses) and $\mathit{\epsilon}$ edges (measurements), the A.C is defined as the second largest Eigen value of the weighted Laplacian $\lambda_2(L_0)$. 2) Average Degree ($\bar d$), indicates the number of edges incident upon its vertices. As $\bar d = \frac{1}{\mathit{v}}\sum_{i=1}^{\mathit{v}}deg(i)$ increases, the number of measurements also increases which eventually improves the MLE estimate. 3) Tree Connectivity, it is shown that WST is directly related to the determinant of the weighted graph Laplacian and the MLE estimate. It is given as $t_w(\mathit{G}) = det(L_w)$, where $L_w$ is the weighted Laplacian.

The above graph connectivity indices provide alternate methods that are computationally less expensive to measure SLAM uncertainty as compared to TOED and IT approaches discussed previously. In Section \ref{SIMULATION RESULTS} we will weigh the effectiveness of our proposed method using these indices.

\section{METHODOLOGY}\label{METHODOLOGY}
The A-SLAM methods outlined in Section \ref{RELATED WORK} assess uncertainty either through the overall map entropy or the SLAM's covariance matrix in its entirety, resulting in high computational costs. In this article, we propose a utility function that incorporates the path entropy and distance to the frontier candidate and adds it to the modern D-Optimality criterion as defined in \cite{NP4}. Computing path entropy is computationally efficient along with penalizing frontiers with large distances. The final utility function not only provides a reliable SLAM estimate but also maximizes the map coverage by minimizing the unknown map area. 

\begin{figure}[thpb]
    \centering
    \includegraphics[height=4.5cm ,width=8.5cm]{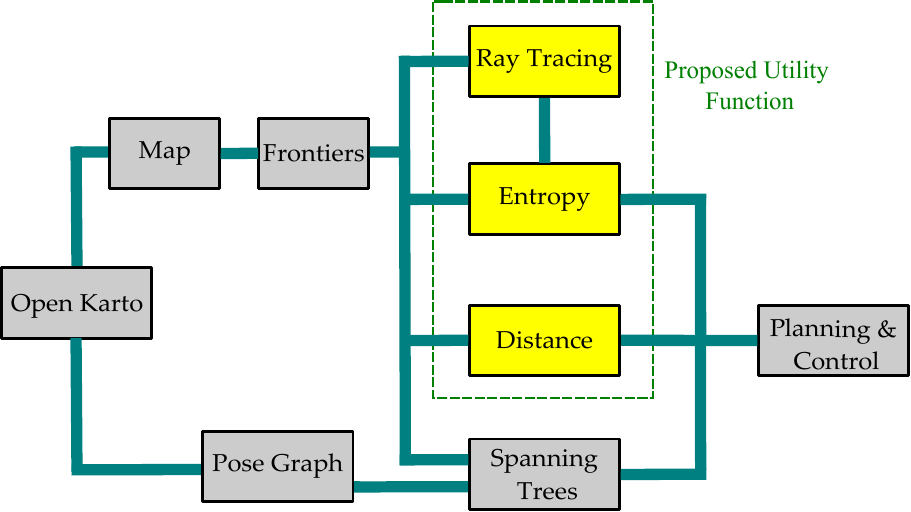}    
    \caption{Framework and proposed utility function}
    \label{fig:1}
\end{figure}

\begin{figure}[thpb]
    \centering
   \includegraphics[width=6cm, height=3cm]{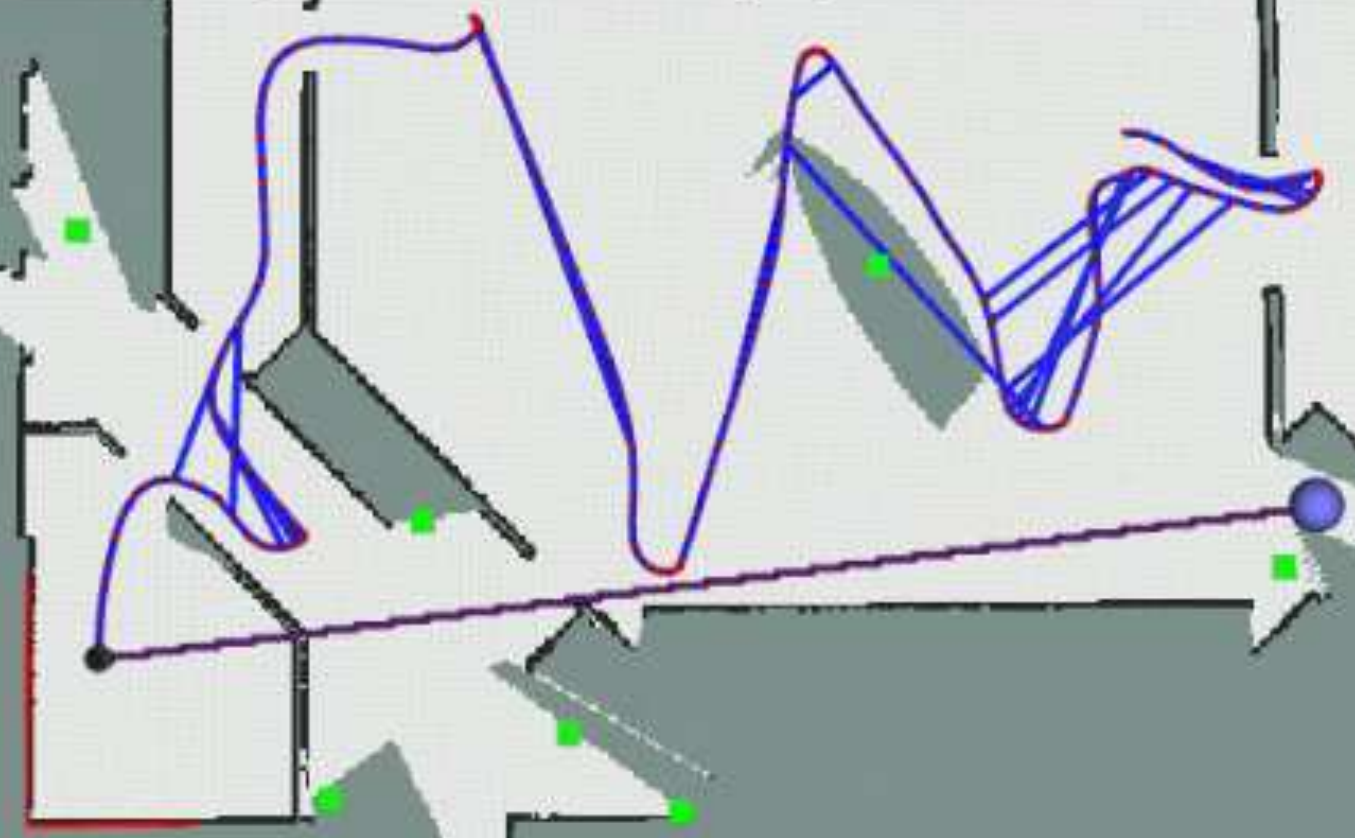}   
    \caption{Proposed utility function ROS implementation, purple line = Bresenham's line, sphere = frontier candidate, Green Squares = frontier centroids. }
    \label{fig:2}
\end{figure}

Figure \ref{fig:1} shows the overview of the proposed utility function. We add this utility function to that of \cite{NP4} which uses Lidar-based Karto SLAM \cite{karto} back-end, with frontier detection over an Occupancy Grid (OG) map. For each frontier centroid (frontier candidate) we apply the Bresenham’s line algorithm \cite{Bresenham} to compute the number of pixels and their occupancy values in a straight line from the robot’s current position. We then compute the path entropy to each frontier candidate and weigh the paths that have a higher number of unknown cells to favor the robot to explore the unknown environment resulting in better coverage.  The lengthy candidate frontier paths are penalized so that the robot does not favor paths longer than a certain threshold and maintains its SLAM accuracy. Finally, the  utility is computed by adding the proposed utility to that of \cite{NP4}. Fig. \ref{fig:2} shows the implementation of the proposed utility function in ROS.

For each candidate frontier $ F = \{f_1, f_2,.....f_N\} \in \mathbb{R}^2 $, we get the occupancy values of each path as $ G^n = \{m_{0}, m_{1} ,.....m_L\}, \forall n \in N$, where $m_{0}...m_{L}$ are the pixel occupancy values of path length $L$. We assign the probability value of $P_{unk} = 0.1$ for unknown pixels (with values = -1) to quantify for low entropy and high information gain (as we are more interested in unknown area of the environment). Occupancy values of obstacles and free space are mapped to probability $P_{Ofree} = 0.45$, weighing high entropy since we are not interested in places already known to the robot. Equation \ref{eq:4} computes the path entropy $E^n$ for each frontier candidate with the above assigned probability values.   

\begin{multline}\label{eq:4}
 E^n = \textit{E}^n[p(m)]_{m \in G^n} = - \sum_{m \in G^n}(P(m_{i,j})log_2(p(m_{i,j})) \\ + P(1- m_{i,j})log_2(1- p(m_{i,j})), \forall m_{i,j} \in M
\end{multline}

Once the path entropy is computed it is normalized with the number of pixels within the path $K^n$,  as shown in Equation \ref{eq:5}. $n=\{n_x, n_y\}$ and $R = \{R_x,R_y\}$ are the selected frontier and robot positions respectively.  

\begin{equation}\label{eq:5}
K^n = \sum_{i=R_x}^{n_x}\sum_{j=R_y}^{n_y}m_{i,j}   
\end{equation}

The path entropy computed in Equation \ref{eq:4} may lead to select frontiers with large distances from the robot. These large distance frontiers may decrease the localization of the robot once it moves toward them.
 For penalizing these frontiers to contain the SLAM uncertainly, we apply an exponential decay operator $\gamma^n$ as shown in Equation \ref{eq:6} and will use it in computing the utility function $U_{2}^{n}$ in Equation \ref{eq:7}. In Equation \ref{eq:6} $\lambda$ is the decay rate operator which acts as a tuning factor for removing frontiers with large distances. We fix $\lambda = 0.6$ since we assume the environment is static and the frontier path entropy remains constant when the robot moves towards the frontier. $n$ and $R$ are frontier and robot locations in $\mathbb{R}^2$ respectively and $dist(*)$ measures the Euclidean distance between $n$ and $R$.  

\begin{equation}\label{eq:6}
\gamma^n = \exp^{-(\lambda*dist(R,n))}
\end{equation}

The utility $U_{2}^{n}$ as shown in Equation \ref{eq:7} is computed by weighing normalized entropy $E^n/K^n$ with  $\rho^n  = 10^{\beta}$, where $\beta$ is a factor which depends on the number spanning trees of the weighted graph Laplacian $(L_w^n)$ computed in Equation \ref{eq:8}. More specifically, $\beta$ is the count of the number of digits before the decimal places of $U_{1}^{n}$ and acts as a balancing factor between entropy and the number of spanning trees.

Finally, we obtain the proposed utility function $U_{tot}$ in Equation \ref{eq:9} as the maximum of the sum  $U_{1}^{n}$ and $U_{2}^{n}$, where Equation \ref{eq:8} is adopted from \cite{NP4}. The advantage of $U_{tot}$ is that it not only provides a good SLAM estimate based on the modern D-Optimality criterion but also increases the coverage of the unknown map by reducing the frontier path entropy and distance.

\begin{equation}\label{eq:7}
U_2^n = (1-E^n/K^n)*\rho^n + \gamma^n
\end{equation}

\begin{equation}\label{eq:8}
U_1^n = \text{Spann}(L_w^n)
\end{equation}

\begin{equation}\label{eq:9}
U_{tot} = \text{max}(U_1^n+U_2^n)
\end{equation}

\section{SIMULATION RESULTS}\label{SIMULATION RESULTS}
The simulations were carried out on ROS Noetic, Ubunto 20.04 on Intel Core i7\textsuperscript{\textregistered}, with a system RAM of 32GB and NVIDIA RTX 1000 GPU. We used the approach of \cite{NP4} and implemented the proposed approach as mentioned in \ref{METHODOLOGY} using Open Karto as SLAM backend, Turtlebot\textsuperscript{\textregistered} equipped with Lidar, Dynamic Window Approach (DWA) \cite{dwa} and Dijkstra's algorithm as local and global planners from the ROS navigation stack. %A video demonstration can %be found here \footnote[2]{\url{https://www.youtube.com/watch?v=XjSBctHOMLY}.}. %
We compared the proposed approach against two different methods which use Frontier Detection based Exploration (FD) \cite{c27} and Active Graph SLAM (AGS) of \cite{NP4} and using two different simulation environments namely the modified Willow Garage (W.G) \footnote[3]{\url{https://github.com/arpg/Gazebo/}.} measuring 2072$m^2$, having no dynamic obstacles, and modified office (Office) measuring 741$m^2$ with obstacles \footnote[4]{\url {https://github.com/mlherd/}}. The ground truth occupancy grid maps were generated using the gazebo\_2Dmap\_plugin \footnote[5]{\url {https://github.com/marinaKollmitz/gazebo}.} which uses wavefront exploration.

For qualitative comparison, we weighed our approach with performance metrics relating to graph connectivity, map efficiency, and \% of the area covered. We used graph connectivity metrics like Algebraic Connectivity (A.C), average degree ($\bar{d}$), normalized tree connectivity ($\hat{\tau}(\mathcal{G})$), and evolution of uncertainty measured as edge D-optimality in the resulting pose-graph. These metrics are strongly related to SLAM estimation accuracy as described in Section \ref{PRELIMINARIES}. Regarding map efficiency matrices, Structural Similarity Index (SSIM) and Root Mean Square Error (RMSE) are used. SSIM$\in (0,1)$ indicates the similarity between two maps by computing the covariance of their pixel values.

We conducted 15 simulations of 30 minutes each for both W.G. and office environments using FD, AGS, and our methods. The results from these simulations along with above mentioned performance matrices are elaborated in Table \ref{table:1} and Fig. \ref{fig:4}.  

	\begin{table}[!h]\scriptsize
		\centering
  		\caption{Average graph connectivity and map quality comparison of 15 simulations (30 minutes each for every method)}	
		\begin{tabular}{|c|c|c|c|c|c|c|}
			\hline
			\textbf{Env.} & \textbf{Meth.}& A.C & $\bar{d}$ & $\hat{\tau}(\mathcal{G})$ & \textbf{SSIM} & RMSE  \\
			\hline
            \multirow{3}{*}{W.G}
			 & FD & 0.104 & \textbf{3.290} & 1.016 & 0.05 & 0.70  \\		
			 & AGS & 0.426 & 2.907 & 1.139 & 0.05  & 0.64   \\		
			 & Our & \textbf{0.845}  & 2.925 & \textbf{1.205}  & \textbf{0.08} & \textbf{0.60}   \\
			\hline
        \multirow{3}{*}{Office}
   		& FD & 3.061 & \textbf{3.179} & 1.229 & 0.09 & 0.83   \\
   	  & AGS & 5.740  & 2.742 & 1.312 &  0.07 & 0.80  \\
	   	  & Our &  \textbf{9.617} & 2.612  & \textbf{1.941} & \textbf{0.11} & \textbf{0.77}  \\
			\hline
		\end{tabular}		
        \label{table:1}
	\end{table}

The first benchmark is FD in which the robot is guided towards the nearest frontier and once it reaches the frontier, it is added to the blacklist of chosen frontiers in order to avoid detecting it again. This approach does not take into account the FIM uncertainty of the pose-graph and nor does it favor revisiting already visited frontiers for loop closing. From Table \ref{table:1} on both environments, we can conclude that this approach provides good $\bar{d}$ because Open Karto creates many loop closure constraints between nodes but does not contribute to uncertainly reduction because the nodes are very close to one another. This method severely lacks in A.C (especially in W.G) as compared to our approach. As described in Section \ref{PRELIMINARIES} A.C is directly related to the accuracy of SLAM estimation and its higher value is encouraged. Further, we observe a lower SSIM and higher RMSE, which renders this approach not suitable for area coverage tasks as compared to preceding methods. Since this method uses a greedy frontier detection search without any quantification of SLAM uncertainly or loop closure, as a result, the SLAM covariance increases, and exploration halts after some time. Eventually the resulting SLAM pose graph has high unreliability and less coverage as compared to Our method. 

The second method in Table \ref{table:1} is AGS. We can infer that this method results in good graph connectivity and map quality matrices in both environments as compared to FD. Especially in office environment the graph connectivity metrics and high. High SSIM and low RMSE indicate better SLAM estimation as compared to W.G. because this environment has more obstacles that bring structure to SLAM estimation.

Our method when compared to the preceding methods in Table \ref{table:1} outperforms them, especially in A.C, $\hat{\tau}(\mathcal{G})$, SSIM, RMSE, and Map size for both environments. We can observe that for Willow Garage our method has 100\% more A.C, 60\% more SSIM, and 6\% less RMSE error than the last best value. For modified Office we get 67\% more A.C, 47\% more $\hat{\tau}(\mathcal{G})$, 22\% more SSIM, and 37\% less RMSE. These promising values indicate the effectiveness of our proposed method.

 \begin{figure}
        \centering
      \subfloat[W.G\label{fig3:1a}]{           \includegraphics[width=8.5cm,height=4cm]{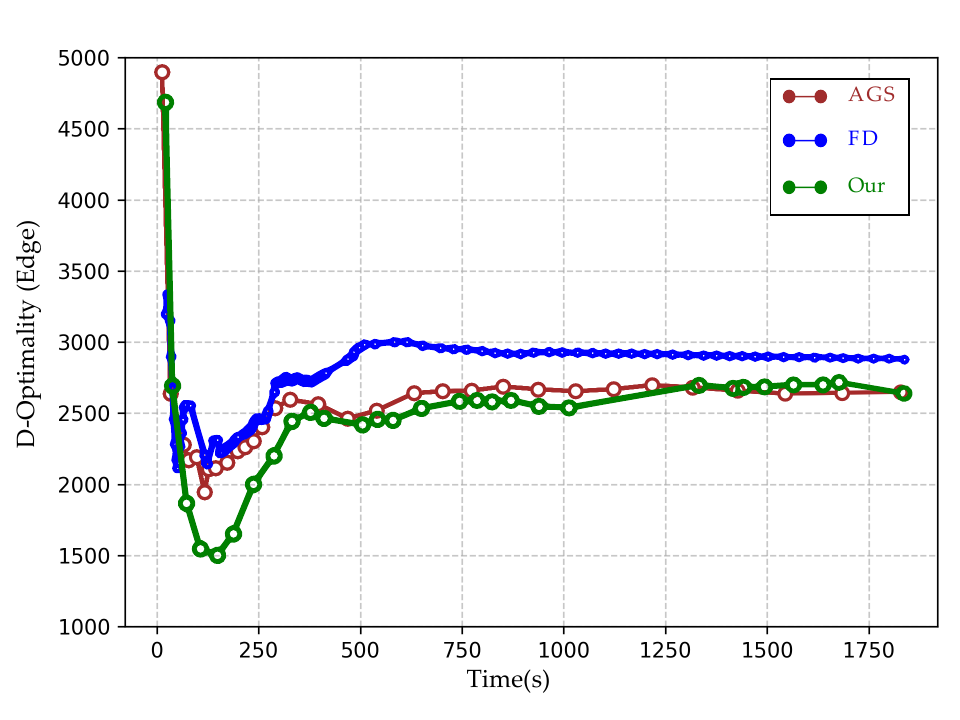}}
        \hfill
      \subfloat[Office\label{fig3:1b}]{%
            \includegraphics[width=8.5cm,height=4cm]{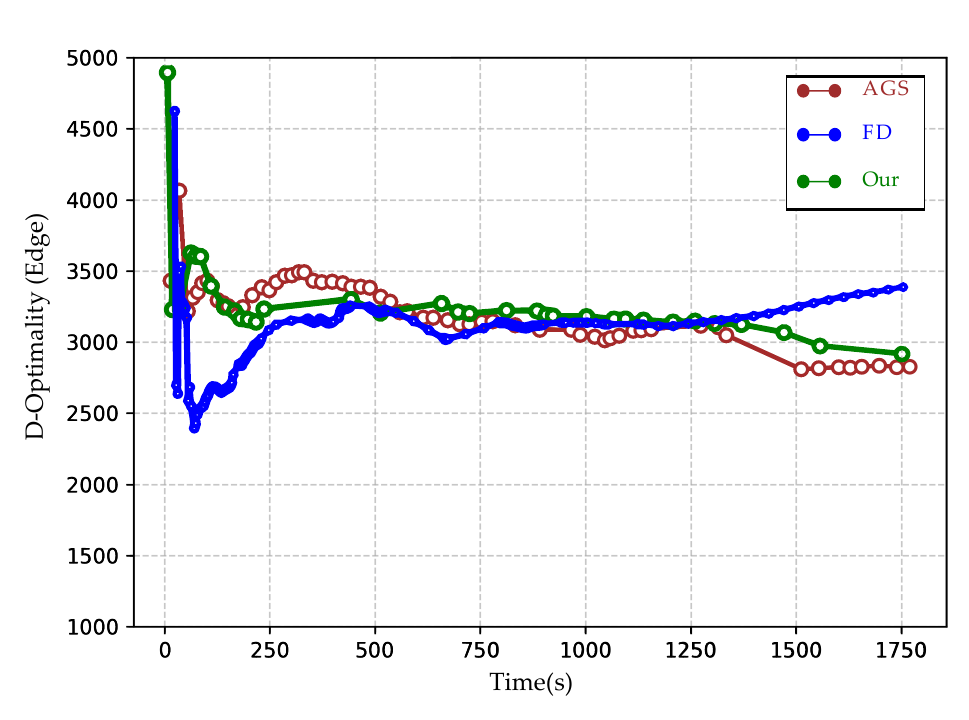}}      
      \caption{Uncertainty evolution of AGS, FD and Our in \ref{fig3:1a} Willow Garage, \ref{fig3:1b} Office Environments.}
      \label{fig:3} 
\end{figure}

	\begin{table}[!h]
		\centering
  		\caption{Uncertainty reduction (\%R) comparison}	
        \label{table:2}
		\begin{tabular}{|c|c|c|c|c|}
			\hline
			\textbf{Env.} & \textbf{Method}& \textbf{D-Opti(Max\&Min)} & Diff & \%R\\
			\hline
            \multirow{3}{*}{W.G}            
			 & FD & 3700\&2900 & 800 & 20 \\	    			
			 & AGS & 4800\&2600 & 2200 & \textbf{45} \\			
			 & Our &  4700\&2600 &  2100  &  44\\
			\hline
   \multirow{3}{*}{Office}
   		& FD & 4600\&3400 & 1200 & 26\\
   	  & AGS & 4100\&2700  & 1400 & 34 \\
	   	  & Our & 4900\&2900  & 2000  & \textbf{40}\\
			\hline
		\end{tabular}	
	\end{table}

\begin{comment}
 \begin{figure}
	\centering
	\begin{minipage}{1\columnwidth}
		\centering
		\includegraphics[width=\textwidth]{results/d_opti_graph_willow.eps}
		\caption{Caption 1}
		\label{label1}
	\end{minipage}%
	\begin{minipage}{1\columnwidth}
		\centering
		\includegraphics[width=\textwidth]{results/d_opti_graph_office.eps}
		\caption{Caption 2}
		\label{label2}
	\end{minipage}
\end{figure}   
\end{comment}

Figure \ref{fig:3} plots the uncertainty evolution over time (s) in the W.G and Office environment. We quantify the uncertainty as D-Optimality for each edge in the entire pose graph. The circles denote the selected goal frontiers. Each method has a different goal frontier detection frequency depending on whether the robot has reached it or not. From Fig. \ref{fig:3} and Table \ref{table:2} we can deduce that initially the uncertainty is high and as the robot explores the environment, it decreases. Our approach and AGS manage to keep the uncertainty bounded to 44\% and 45\% of their maximum threshold respectively, while FD keeps it at 20\% hence resulting in a poor SLAM estimate due to lack of loop closure and using a greedy frontier search. 

    \begin{figure}[thbt!] 
        \centering
      \subfloat[W.G\label{fig4:1a}]{%
           \includegraphics[width=7.5cm,height=4cm]{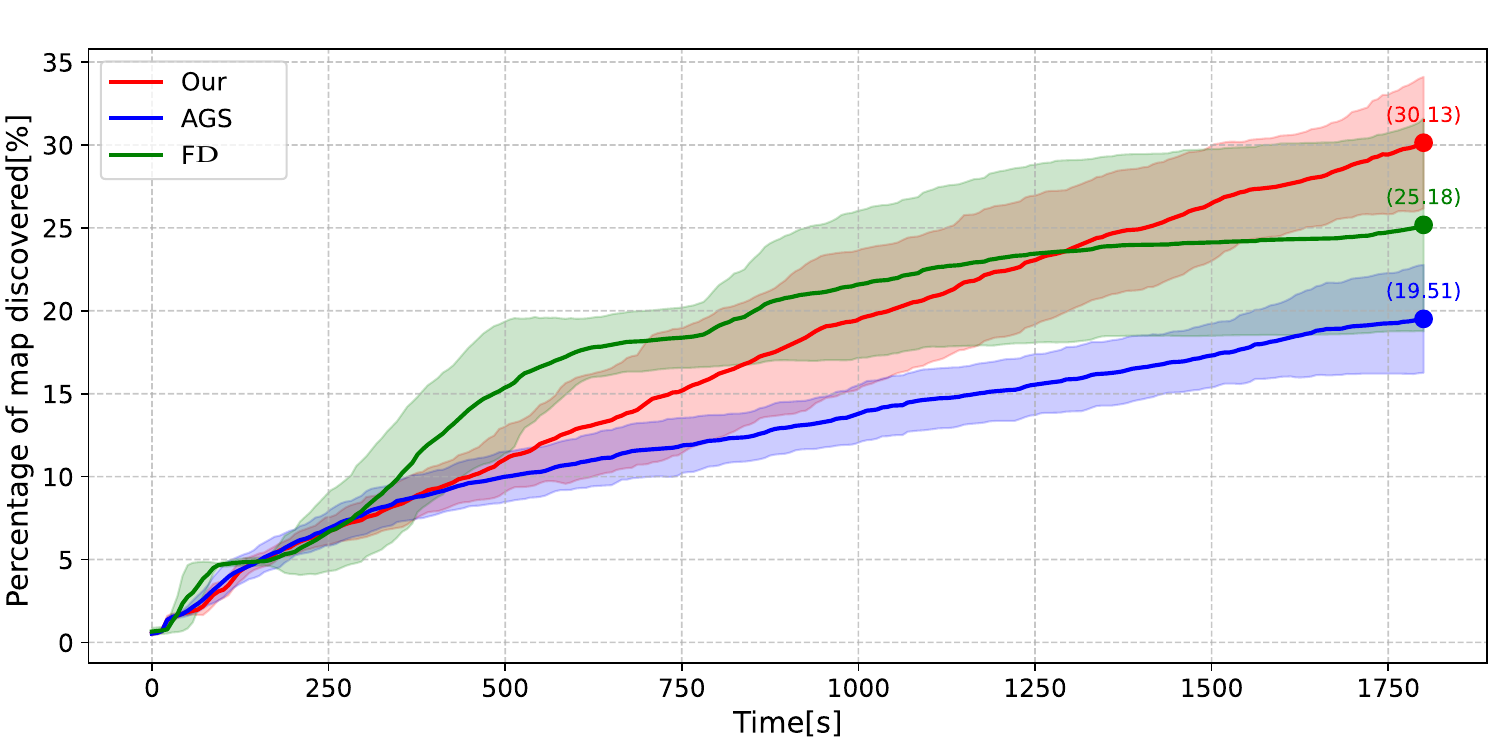}}
        \hfill
      \subfloat[Office\label{fig4:1b}]{%
            \includegraphics[width=7.5cm,height=4cm]{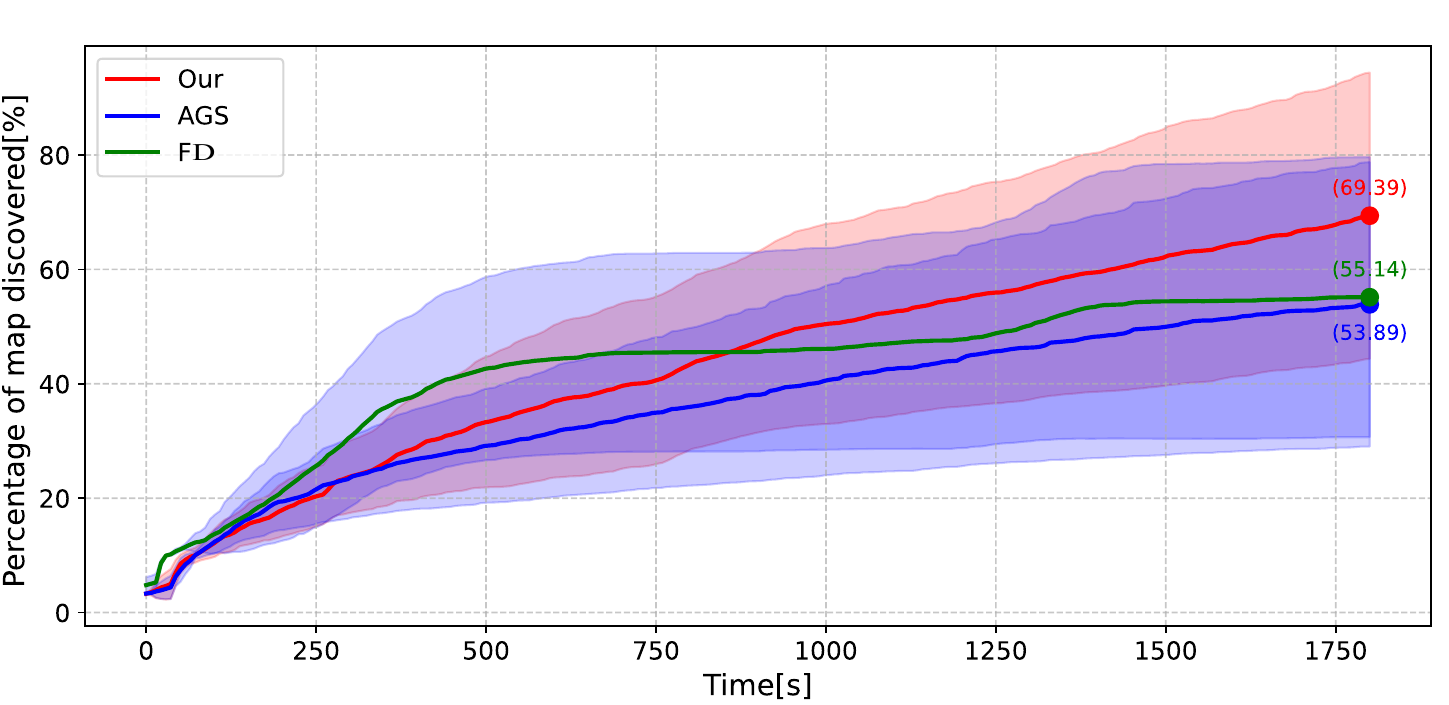}}  
            
      \caption{Comparison of evolution of map discovered with average and standard deviation for our, AGS and FD methods. With 15 simulations (30 minutes each for every method) }
      \label{fig:4} 
    \end{figure}

\begin{figure}[thbt!]  
    \centering
    \subfloat[W.G\label{fig5:1a}]{%
       \includegraphics[width=7cm,height=5cm]{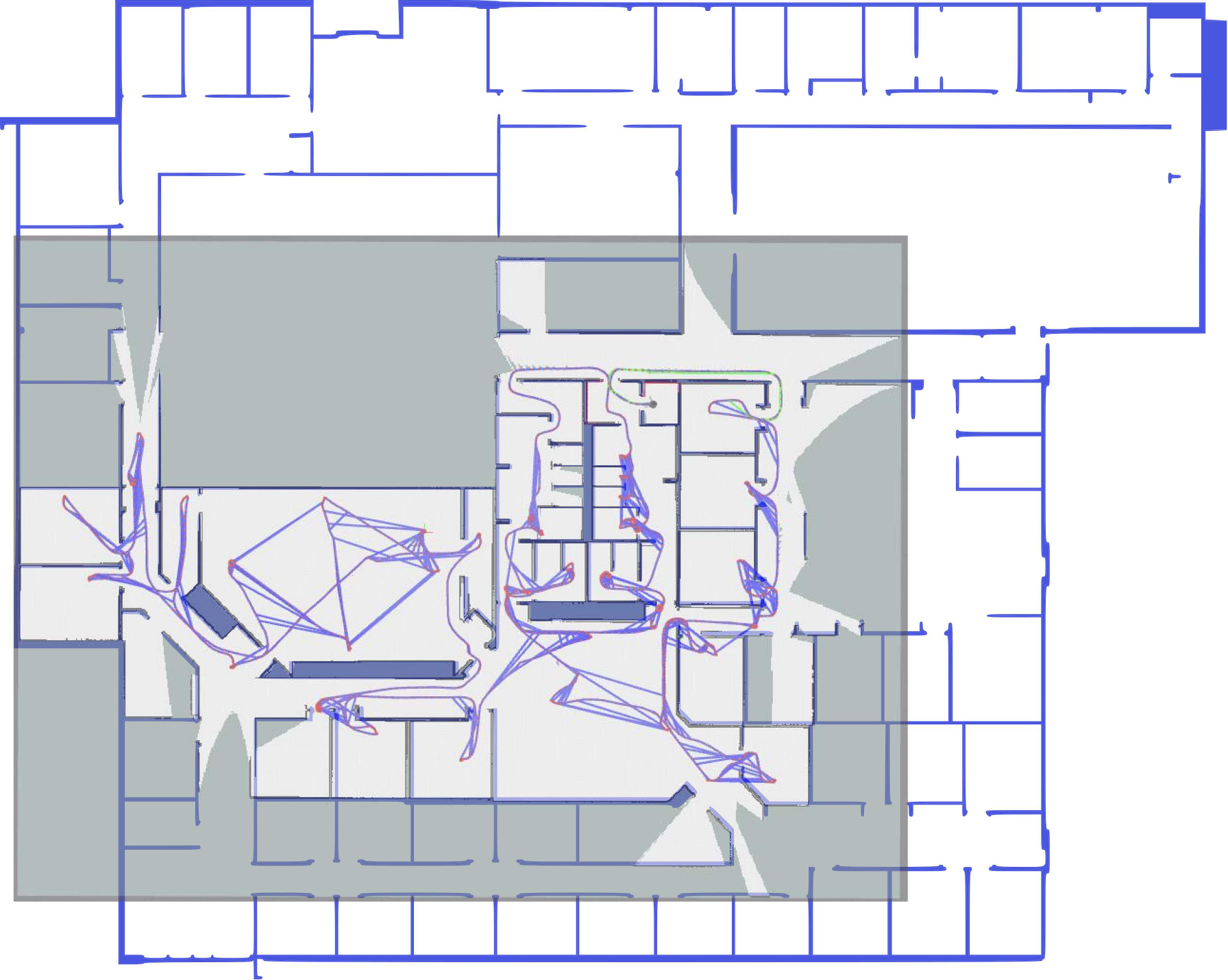}}
    \hfill
    \subfloat[Office\label{fig5:1b}]{%
        \includegraphics[width=7cm,height=4cm]{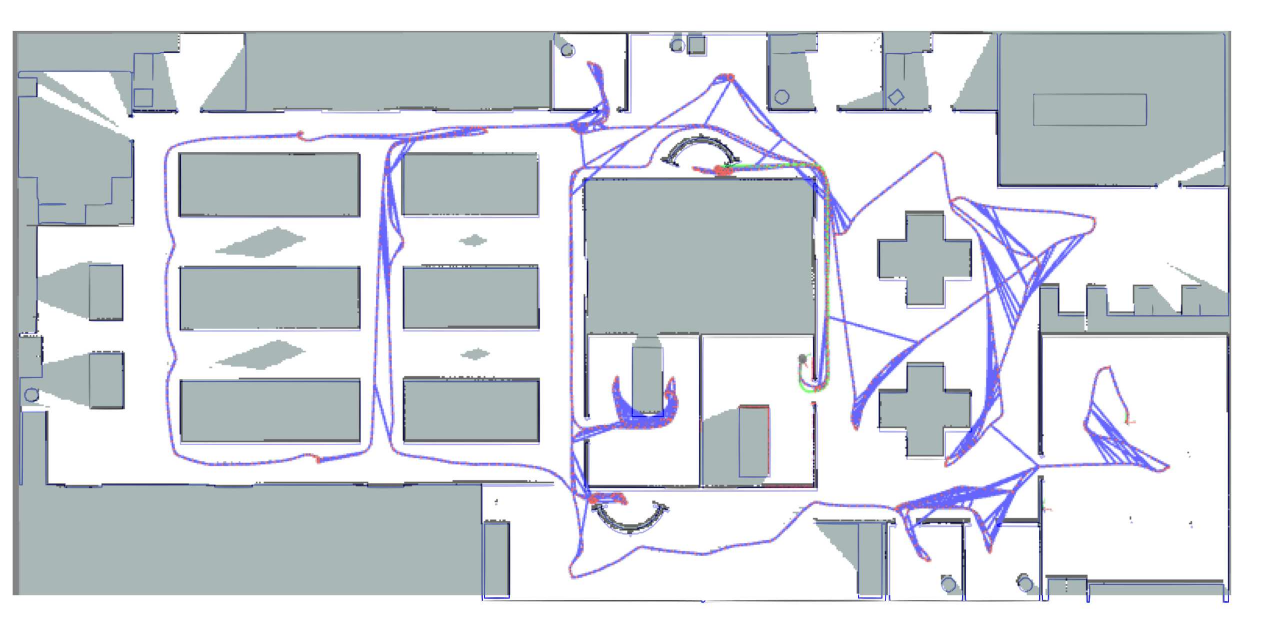}}    
    \caption{Obtained pose graphs using Our approach and ground truth maps (blue).}
    \label{fig:5} 
    \end{figure}

In Fig. \ref{fig:4} a comparison of the evolution of the percentage of map discovered is presented. The average values along with the standard deviation of 15 simulations, 30 min each of FD, AGS, and our are shown. Since FD uses a greedy frontier detection and exploration approach it starts with a higher slope than AGS and our method till 1000 and 500 seconds for W.G and Office respectively, after that it the SLAM covariance becomes unbounded, and the slope decreases with the final percentage of discovered map at 25.18\% and 55.14\% respectively. Both AGS and our method manage to keep the slope of average exploration values constant but our method eventually manages to explore 54\%, 30\% more area than AGS, and 20\%, 25\% more area than FD for both W.G and Office environments respectively. In Fig. \ref{fig:5} the resulting occupancy grid maps and pose graphs are overlapped on the ground truth maps to show the area covered by our approach in environments.

\section{EXPERIMENTAL RESULTS}\label{EXPERIMENTAL RESULTS}
\begin{figure}[H] 
    \centering
      \subfloat[RosBot 2\label{fig6:1a}]{%
           \includegraphics[width=0.3\linewidth]{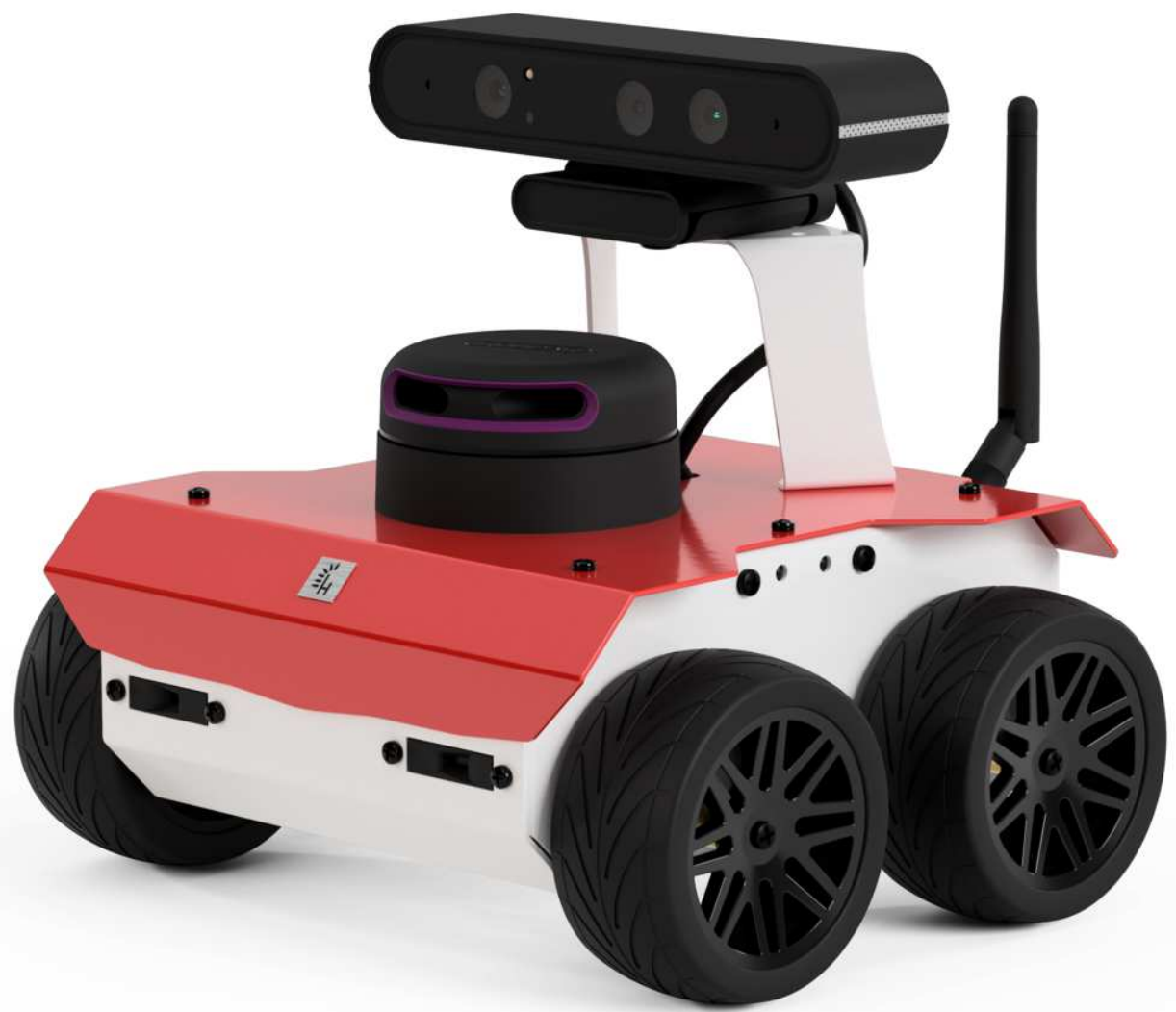}}
        \hfill
      \subfloat[Mapped environment\label{fig6:1b}]{%
            \includegraphics[width=0.65\linewidth]{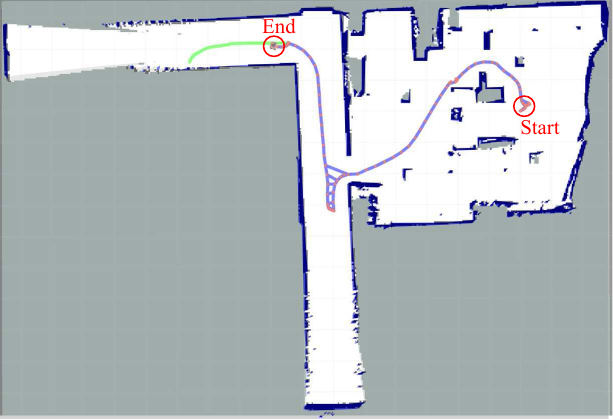}}        
      \caption{Robot used for experiments and the resulting mapped environment overlapped with ground truth map(blue).}
      \label{fig:6} 
    \end{figure}

The experiments were performed using four-wheel diff. drive ROSBot2 robot\footnote[3]{\url{https://husarion.com/manuals/rosbot/} .}as shown in Fig. \ref{fig6:1a} and ROS (noetic) on Ubunto 20.04.6 (LTS) running on Intel Xeon\textsuperscript{\textregistered} W-2235 CPU 3.80GHz x 12, 64Gb RAM and Nvidia Quadro RTX 4000 GPU.  

The environment consists of a room (lab environment) with static obstacles and two corridors measuring 81$m^2$ in total. We used mapping efficiency and exploration time as performance matrices for experimental results. Fig. \ref{fig6:1b} shows the computed OG map, and SLAM pose graph along with robot start and end positions using our approach. The computed OG map by the robot is overlapped with that of the ground truth (blue). We can observe high similarity between the two maps, rendering the high mapping efficiency of our approach. 

\begin{figure}[H]
    \centering
    \includegraphics[width=8cm, height=4cm]{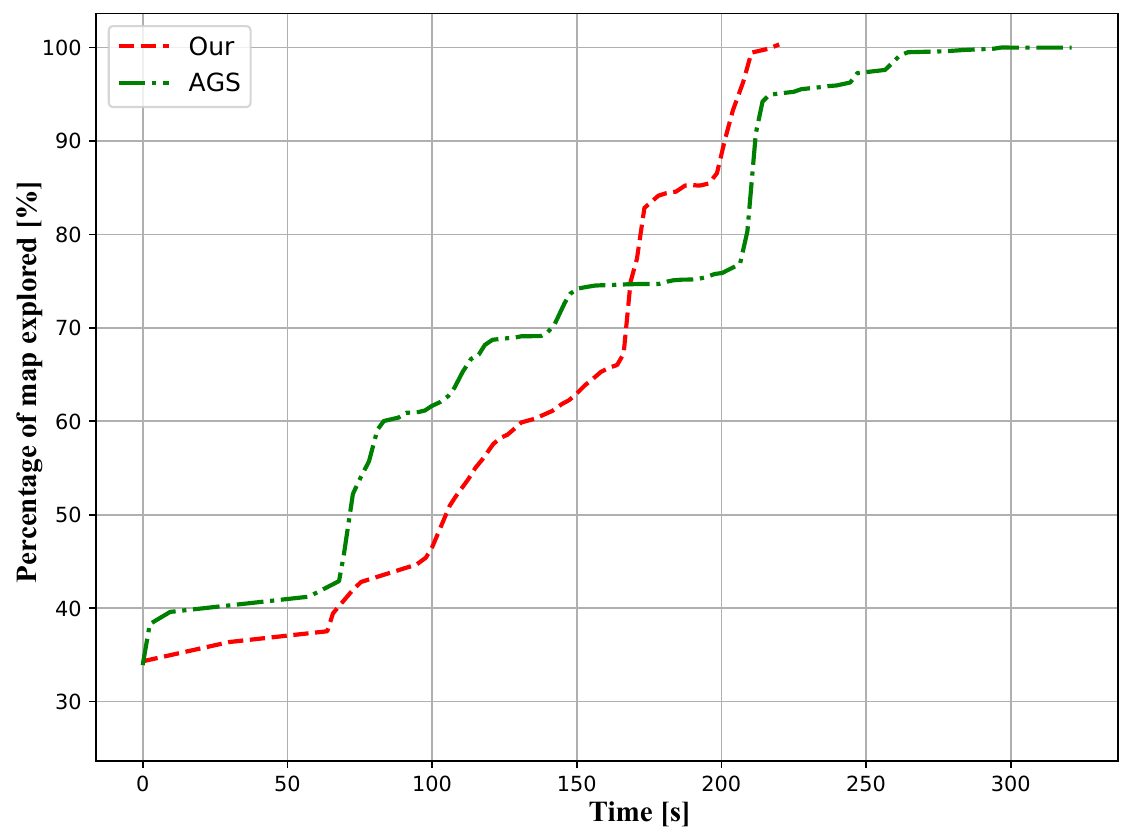}   
    \caption{Average Coverage Percentage in four experiments.}
    \label{fig:7}
\end{figure}

Figure \ref{fig:7} shows the average percentage of four experiments (2-AGS, 2-Our) for a total duration of 350 seconds. We can observe that initially both AGS and our method approximately map 35\% of the environment this is because of the high range (up to 16m) of the Lidar sensor \footnote[4]{\url{https://www.slamtec.com/en/Lidar/A2/} .} of RosBot 2 robot. With the evolution of time we can observe that using our utility function, a steep slope is obtained from 60 to 210 seconds (approx.) which helps the robot to cover the entire environment in 230 seconds as compared to 280 seconds for AGS. These results are in contrast with Section \ref{SIMULATION RESULTS} and indicate that our approach supersedes AGS and therefore can be utilized for efficient exploration of the environment while maintaining good SLAM efficiency. 

\section{CONCLUSIONS}\label{CONCLUSIONS}

In this article, we have presented a utility function that selects the most favorable frontier goal location within an occupancy grid map for a reliable A-SLAM with an area coverage task. The proposed utility function incorporates path entropy to select the frontier goal location which has the highest amount of unknown cells within its path thus maximizing the area coverage. Using simulation and experimental results on publicly available environment maps we have proved the efficiency of our approach as compared to similar methods. As a future prospective, we plan to incorporate our method in a multi-robot scenario utilizing efficient frontier-sharing for maximum environment exploration.

\section*{Acknowledgment}
This work was conducted within the framework of the NExT Senior Talent Chair DeepCoSLAM, funded by the French Government through the program "Investments for the Future" administered by the National Agency for Research (ANR-16-IDEX-0007). We also extend our gratitude to the Région Pays de la Loire and Nantes Métropole for their invaluable support in facilitating this research endeavour.

\end{document}